\documentclass[11pt,letterpaper]{article}
\usepackage[margin=1in,textwidth=5.5in]{geometry}
\usepackage[T1]{fontenc}
\usepackage{textcomp}
\usepackage{times}
\usepackage{natbib}
\setcitestyle{authoryear,round,citesep={;},aysep={,},yysep={;}}
\usepackage{primeseeker_boxes}

\usepackage{amsmath,amsfonts,bm}

\def\eqref#1{equation~\ref{#1}}

\def\1{\bm{1}}

\DeclareMathAlphabet{\mathsfit}{\encodingdefault}{\sfdefault}{m}{sl}
\SetMathAlphabet{\mathsfit}{bold}{\encodingdefault}{\sfdefault}{bx}{n}

\usepackage{hyperref}
\hypersetup{%
  pdftitle={PrimeSeeker: Capability-Oriented Supervision for Deep Search Agents},
  pdfauthor={Linzhi Peng, Hanting Chen, Heng Chang, Ke Cheng, Bowen Du, Weifeng Lv}%
}
\usepackage{url}
\usepackage{graphicx}
\usepackage{subcaption}
\usepackage{booktabs}
\usepackage{array}
\usepackage{tabularx}
\newcolumntype{Y}{>{\centering\arraybackslash}X}
\newcommand{\best}[1]{\underline{#1}}
\title{PrimeSeeker: Capability-Oriented Supervision for Deep Search Agents}
\author{%
Linzhi Peng\textsuperscript{1} \quad
Hanting Chen\textsuperscript{2} \thanks{Corresponding author}\quad
Heng Chang\textsuperscript{3}\thanks{Project organizer}\\[0.3em]
Ke Cheng\textsuperscript{1} \quad
Bowen Du\textsuperscript{1} \quad
Weifeng Lv\textsuperscript{1}\\[0.8em]
{\small \textsuperscript{1}Beihang University},
{\small \textsuperscript{2}Huawei Technologies Ltd.},
{\small \textsuperscript{3}Tsinghua University}\\[0.5em]
{\small \texttt{lzpeng626@buaa.edu.cn, hantingchen@pku.edu.cn}}\\
{\small \texttt{changh17@tsinghua.org.cn, ckpassenger@buaa.edu.cn}}\\
{\small \texttt{dubowen@buaa.edu.cn, weifenglv@buaa.edu.cn}}%
}

\begin{document}

\maketitle

\begin{abstract}
Large language model search agents are often trained with synthetic questions whose difficulty is increased through larger evidence graphs, additional hops, and longer trajectories. These global properties, however, are only indirect proxies for the local retrieval capabilities required during search. To address this mismatch, we introduce latent anchor reasoning, which consists of resolving an unnamed retrieval anchor from descriptive specifications and transferring the recovered anchor into a subsequent information demand. 
This primitive retrieval unit decomposes deep search into chains of coupled operations and organizes question construction around anchor resolution and relation transfer, without prescribing a canonical search path.
Based on this formulation, we propose PrimeSeeker, a capability-oriented framework that constructs web-grounded anchor structures and jointly derives a question and a reference evidence skeleton.
The skeleton preserves supporting evidence from construction and guides
expert generation through extractive highlights of current tool
observations. These highlights are removed before supervised fine-tuning,
while the skeleton is subsequently reused to audit reference-step
coverage for reinforcement-learning rewards.
We construct 9,221 expert trajectories, training a 30B search agent. Across five deep-search benchmarks, PrimeSeeker achieves strong performance, while reference-step optimization further improves the supervised policy. The resulting trajectories exhibit low retrieval redundancy, and fixed-budget evaluation shows strong solution coverage with substantially fewer tool calls than long-horizon systems. Data and code are available at  https://github.com/PengLinzhi/PrimeSeeker.
\end{abstract}

\section{Introduction}
\label{sec:introduction}

Large language models (LLMs) are increasingly used as tool-augmented agents
that interleave reasoning with external retrieval to solve knowledge-intensive
tasks
\citep{NEURIPS2020_6b493230,yao2023react}. Recent systems acquire such behavior
through iterative-retrieval supervision, outcome-based reinforcement learning,
and multi-stage agentic training
\citep{NEURIPS2025_566aad4f,jin2025searchr,
song2025r1searcherincentivizingsearchcapability,tongyi_deepresearch}.
Although these approaches have substantially advanced long-horizon information
seeking, they leave open what specific search capabilities a constructed
question and its expert trajectory are intended to teach.

Most data construction pipelines model search capability through two indirect routes. The first increases the difficulty of the task by expanding the evidence graphs, obscuring entities, and dispersing the supporting evidence
\citep{li2025websailornavigatingsuperhumanreasoning,websailor_v2, openseeker,redsearcher}. The second removes questions that can be solved through simple shortcuts or simplifies trajectories by pruning redundant evidence.
\citep{fort_searcher,wang-etal-2026-webclipper}.
By improving task difficulty or trajectory, they refine the surface properties of training examples. Yet they fall short of specifying the search capability—a reusable behavioral operation invariant across different content instances.
A complex question may be quickly solved using parametric knowledge or a single discriminative query, while a short question may still require many steps and ultimately demand identifying an unnamed target. 
The same retrieval ability may also appear in questions of different complexity and be expressed through different search paths. Therefore, complexity and trajectory properties alone cannot specify the capability being supervised.

\begin{figure}[t]
    \centering
    \includegraphics[width=0.8\linewidth]{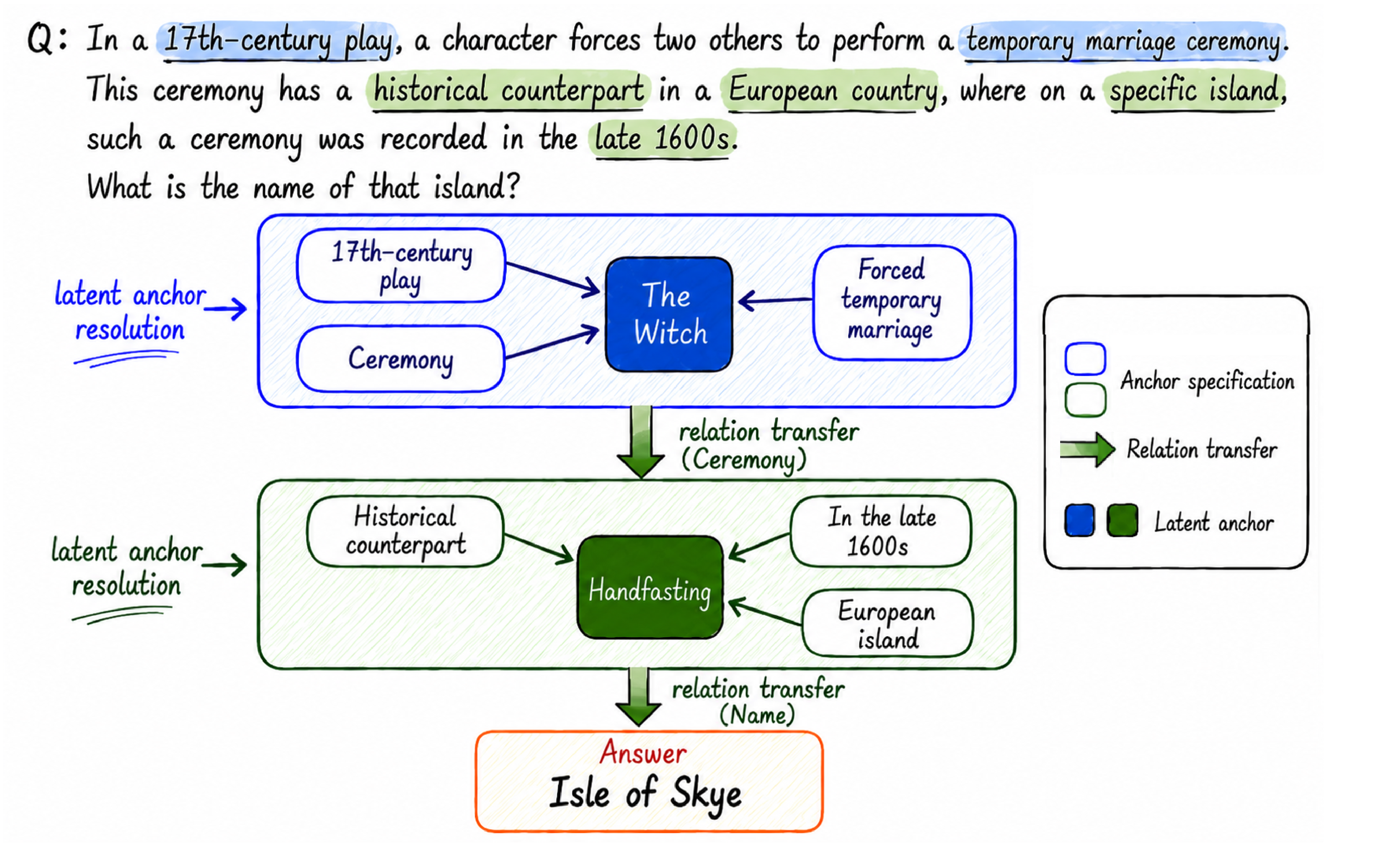}
    \caption{
    Illustration of latent anchor reasoning. Descriptive specifications first
    identify an unnamed play; the recovered anchor makes the ceremony relation
    queryable, leading to a second latent anchor and ultimately the name of the requested
    location. The decomposition specifies local retrieval capabilities without
    prescribing a unique search trajectory.
    }
    \label{fig:latent-anchor-reasoning}
\end{figure}

We address this mismatch by viewing deep search as a composition of local
retrieval capabilities. In many web search questions, the next useful target
is not explicit and is specified only through descriptive clues, which we call
\emph{anchor specifications}. The agent must first recover a usable retrieval
anchor from these specifications and then use the recovered anchor to formulate
a subsequent information demand. We refer to these coupled operations as
\emph{latent anchor resolution} and \emph{relation transfer}, and to the overall
capability as \emph{latent anchor reasoning}. Figure~\ref{fig:latent-anchor-reasoning}
illustrates how multiple such local units compose a deep search problem without
determining the model's realized search path. This decomposition defines the targeted retrieval capabilities without enforcing a canonical question or search path.

We propose \textsc{PrimeSeeker}, a capability-oriented framework that constructs
questions and expert trajectories around these local operations. PrimeSeeker
builds web-grounded anchor structures by collecting a small number of
independently sourced and validated specifications for each anchor and following
substantive semantic relations. From the same construction structure, it derives
both a question \(q\) and a reference evidence skeleton \(D_q\). During expert
generation, \(D_q\) highlights matching evidence only after it appears in the
current tool observation, without introducing unseen information or determining
the expert's next action. All skeleton-dependent guidance is removed before
supervised fine-tuning, while the same skeleton remains external to the policy
and can subsequently audit reference-step coverage during reinforcement
learning.

Using this pipeline, we construct 9,221 expert trajectories and train
PrimeSeeker from Qwen3-30B-A3B-Thinking-2507. Across two expert models,
reference-aligned \(D_q\) guidance consistently improves expert accuracy,
whereas random observation-derived highlights sharply degrade performance.
The generated demonstrations exhibit lower duplicate-search and
duplicate-visit rates than the compared search-agent corpora. Across five deep search benchmarks, PrimeSeeker-\(D_q\) consistently
improves over PrimeSeeker-SFT by reusing the construction-derived skeleton for
reference-step coverage. Under a controlled interaction budget, PrimeSeeker
also matches or exceeds strong long-horizon systems while using substantially
fewer tool calls. These results demonstrate the utility of reusing construction-derived
evidence across expert generation and policy optimization.


Our contributions are summarized as follows:
\begin{itemize}
    \item We reformulate deep search from a capability-oriented perspective and
    introduce latent anchor reasoning, characterizing each question
    through latent anchor resolution and relation transfer. This enables an operational basis for organizing search-task construction.

    \item We propose PrimeSeeker, which jointly derives a question and a
    reference evidence skeleton \(D_q\) from the same anchor structure. The skeleton guides expert generation and supplies reference-step rewards, providing consistent supervision from task construction to policy optimization while preserving diverse valid search strategies.

    \item We evaluate PrimeSeeker across five deep search benchmarks. The resulting demonstrations exhibit low
    retrieval redundancy, \(D_q\)-guided generation consistently improves expert
    accuracy, reference-step optimization improves the supervised agent, and
    fixed-budget evaluation shows strong solution coverage with substantially
    lower tool consumption.
\end{itemize}

\section{Problem Formulation}
\label{sec:problem-formulation}

We study deep search tasks that require an agent to interleave reasoning with
multi-turn retrieval. Given a question \(q\), a retrieval environment
\(\Sigma\), and a gold answer \(y^\star\), the agent must acquire external
evidence and use it to support the final answer. Instead of treating the length
or global structure of a complete trajectory as the basic unit of supervision,
we focus on the local retrieval capabilities from which such trajectories are
composed.
We represent the task-relevant evidence structure behind \(q\) as
\[
\mathcal{G}_q
=
(\mathcal{A}_q,\mathcal{C}_q,\mathcal{R}_q,\mathcal{E}_q,y^\star).
\]
Here, \(\mathcal{A}_q\) contains the task-relevant retrieval anchors. Anchors may be named entities, rules, attribute slots, or
other intermediate objects that enable subsequent retrieval.
\(\mathcal{C}_q\) collects the descriptive specification sets \(C(a_i)\)
associated with these anchors, \(\mathcal{R}_q\) contains the queryable relation
demands over them, and \(\mathcal{E}_q\) contains the external webpages
supporting the specifications and relations.

For an anchor \(a_i\in\mathcal{A}_q\), let
\[
C(a_i)=\{c_{i1},\ldots,c_{ik}\}
\]
denote its specification set. Each \(c_{ij}\) describes a property, function,
scenario, context, or partial relation of \(a_i\), without necessarily exposing
a directly searchable name. Therefore, the specification induces a candidate
space
\[
\operatorname{Cand}_{\Sigma}(C(a_i))
\]
within the retrieval environment.
We write the primitive local retrieval unit as
\[
C(a_i)\Rightarrow a_i\xrightarrow{r_i}z_i.
\]

The first operation, \(C(a_i)\Rightarrow a_i\), is \emph{latent anchor
resolution}. The agent forms candidate anchors from the specifications,
retrieves discriminative evidence, and identifies the corresponding explicit
anchor. Resolution may be immediate when a strong prior or discriminative query
sharply narrows the candidate space, or may require iterative comparison and
verification. If the anchor is already \emph{explicit}, \(C(a_i)\) degenerates to \(\{a_i\}\).

The second operation, \(a_i\xrightarrow{r_i}z_i\), is \emph{relation transfer}.
Once \(a_i\) is available, the agent formulates an information demand
conditioned on that anchor. The result takes one of three forms:
\[
z_i\in\{y^\star,\ a_{i+1},\ C(a_{i+1})\}.
\]
It may directly yield the final answer \(y^\star\), expose the explicit anchor
\(a_{i+1}\), or produce a set of specifications
\(C(a_{i+1})\) that defines a new latent anchor. 
Semantic migration occurs when relation transfer moves the search to \(z_i\), thereby composing multiple local retrieval units into a deep search task.

By specifying capabilities rather than execution routes, the formulation
allows the same anchor to be recovered through prior knowledge, a
discriminative query, or iterative evidence comparison. Supervision therefore
targets identifiable instances of latent anchor resolution and relation
transfer, together with their grounding in external evidence, rather than
adherence to a fixed search path.
\section{Method}
\label{sec:method}

\begin{figure}[t]
    \centering
    \includegraphics[
        width=\linewidth,
        keepaspectratio
    ]{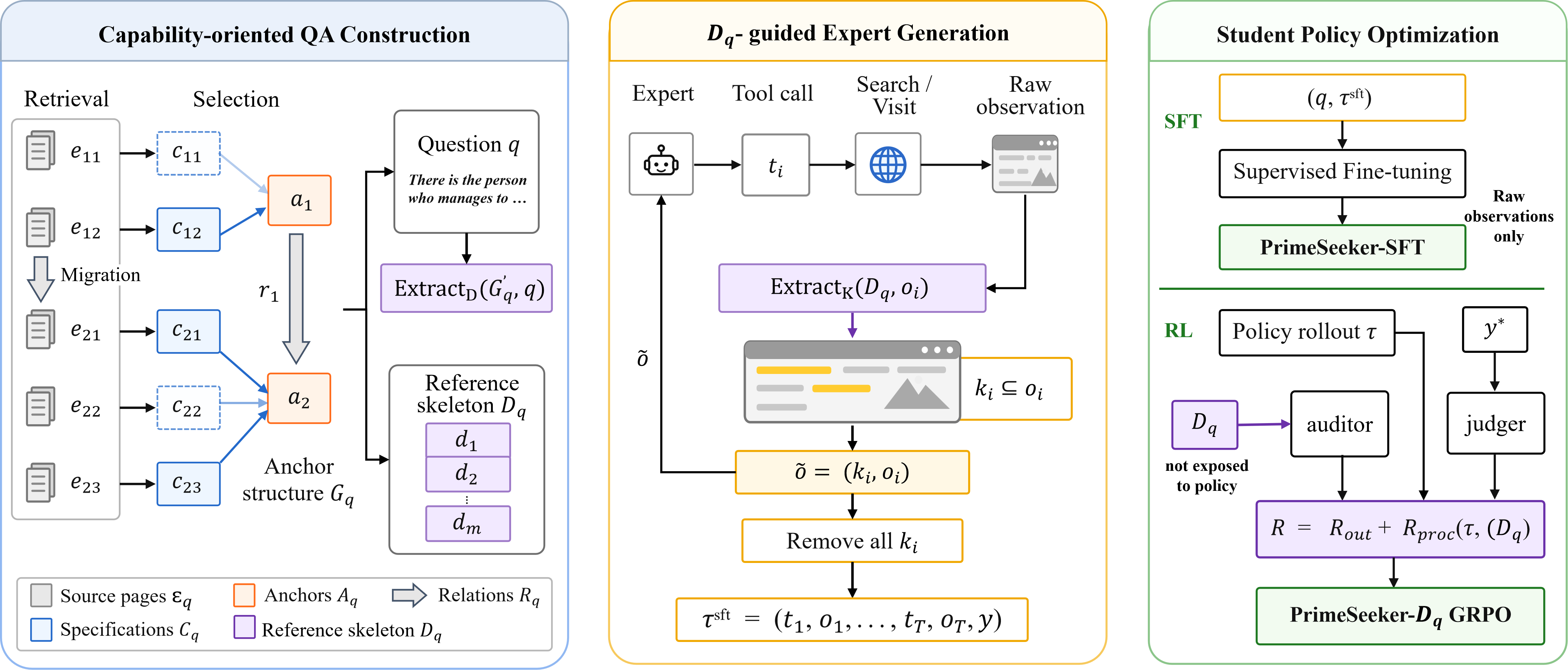}
    \caption{
    Overview of \textsc{PrimeSeeker}. The constructed anchor structure yields both
    a question \(q\) and a reference evidence skeleton \(D_q\). During expert
    generation, \(D_q\) extracts matching evidence from the current observation.
    These annotations are removed before SFT, while \(D_q\) remains external to the
    policy and later provides reference-step coverage signals during reinforcement
    learning.
    }
    \label{fig:method-overview}
\end{figure}

PrimeSeeker operationalizes the capability-oriented formulation in
Section~\ref{sec:problem-formulation} through capability-oriented QA construction and \(D_q\)-guided expert generation.
The recovered trajectories are used for supervised policy training, while the
evidence skeleton \(D_q\) can optionally be reused to audit
reference-step coverage during reinforcement learning.
Figure~\ref{fig:method-overview} presents the complete pipeline.

\subsection{Capability-oriented QA Construction}
\label{sec:qa-construction}

PrimeSeeker constructs questions around explicit instances of latent anchor
resolution and relation transfer, rather than increasing global question
complexity and relying on the resulting trajectory to expose useful search
behavior.

\paragraph{Constructing the anchor structure.}
Starting from a seed page or topic, the construction agent expands the current
anchor \(a_i\) by retrieving independent webpages. It extracts one factual
specification \(c_{ij}\) from each source page
\(e_{ij}\in\mathcal{E}_q\) and retains the corresponding provenance, so that
every specification can be traced to supporting evidence.

A language model validator retains a specification only if it is supported by
the source page, adds information not already covered by the collected
specifications, and remains relevant to the relation chain. Through
ambiguity checking validation, we ensure that the retained
anchor is unique and the specifications are correct. We collect two or
three specifications for each anchor from different webpages. The combined
description is then checked for ambiguity and retained only when it is
sufficiently discriminative to identify a single anchor in practice.

The construction agent then follows a substantive semantic relation
\[
a_i\xrightarrow{r_i}a_{i+1},
\]
such as a functional, historical, causal, compositional, or institutional relation. Iterative descriptive specification collection and semantic
migration produce the construction record
\[
\mathcal{G}'_q
=
(\mathcal{A}_q,\mathcal{C}_q,\mathcal{R}_q),
\]
together with the source provenance associated with each specification.

\paragraph{Question and reference skeleton derivation.}
The question generator selects a connected substructure of
\(\mathcal{G}'_q\) and verbalizes its specifications and relation demands.
For anchors intended to train latent anchor resolution, the question includes
only their descriptive specifications, without revealing their identities.
The selected relation demands are retained so that each recovered anchor can
support the intended subsequent query. The resulting substructure is then
rendered as a natural language question \(q\) with target answer \(y^\star\).

From the same construction record, PrimeSeeker extracts a reference evidence
skeleton
\[
D_q
=
\operatorname{Extract}_{D}(q,\mathcal{G}'_q)
=
\{d_1,\ldots,d_m\}.
\]
Each reference step \(d_i\) links a supporting fact to the anchor or relation established by that fact.
Together, the steps describe the evidence milestones that a successful search
is expected to establish, progressing from intermediate anchors and relations
to evidence for the target answer. They specify what evidence is relevant
without prescribing the queries, actions, or order used to retrieve it.
A reference step is treated as answer evidence when the anchor or relation it
verifies directly establishes \(y^\star\) and the remaining steps capture
intermediate evidence progress.

\paragraph{Quality control.}
We remove candidates that the designated expert answers correctly without
external tools under the closed-book filtering configuration. This excludes
questions already solved by that expert from parametric knowledge alone.

\subsection{\texorpdfstring{\(D_q\)}{Dq}-guided Expert Generation}
\label{sec:trajectory-generation}

During expert generation, PrimeSeeker temporarily highlights relevant evidence
only after it appears in the current observation, while excluding spans that
directly reveal the target answer. All auxiliary guidance is removed before
SFT, so the student is trained solely on the original interaction history.

\paragraph{Evidence extraction and guidance.}
At turn \(i\), the expert issues a tool call \(t_i\) and receives a raw
observation \(o_i\). PrimeSeeker extracts evidence from the current observation
that is relevant to the question and its reference steps:
\[
k_i
=
\operatorname{Extract}_{k}(q,o_i,D_q),
\qquad
\tilde{o}_i=(o_i,k_i).
\]
The guidance is strictly extractive:
\[
k_i\subseteq o_i.
\]
Every highlighted span must occur verbatim in the current raw observation. If
no relevant evidence is found, \(k_i\) is empty and \(o_i\) is left unchanged. 
Spans that directly reveal the target answer \(y^\star\) are
excluded, and the reference skeleton itself is never inserted into the observation. 
The expert remains responsible for forming hypotheses, selecting
tools, and choosing the search path, without any objective on trajectory length
or repetition. The guidance therefore directs attention to relevant evidence
without constraining how the answer is reached.

\paragraph{Guidance removal and consistency filtering.}
The expert is instructed not to reference guidance markers or hidden
annotations in its reasoning. After generation, all \(D_q\)-dependent
annotations and extracted highlights are removed, yielding
\[
\tau^{\mathrm{sft}}
=
(t_1,o_1,\ldots,t_T,o_T,\hat{y}_{\tau}),
\]
where free-form assistant reasoning is omitted from the notation.

A prompt-based verifier then checks whether the retained reasoning and tool
calls are supportable from \(q\) and the raw interaction history. Trajectories
that explicitly depend on hidden annotations or contain actions unsupported by
the available history are discarded. Removing the guidance ensures that no
additional factual content appears in the student input, although the expert
policy may still benefit from evidence highlighting during synthesis.

\subsection{Student Policy Optimization}
\label{sec:training}

\paragraph{Supervised fine-tuning.}
We train the student on the recovered trajectories using standard next-token
prediction over assistant reasoning, tool calls, and final answers. Raw tool
observations remain in the context but are masked from the loss. The student
therefore receives neither the reference evidence skeleton \(D_q\) nor the
extracted highlights \(k_i\).

\paragraph{Reference-step coverage reward.}
A reward based only on final correctness cannot distinguish an unsuccessful
rollout that has recovered useful evidence from one that has made no relevant
progress. We therefore use \(D_q\) during reward computation to measure how
much of the reference evidence skeleton is established by the rollout, while
keeping \(D_q\) hidden from the policy.

Given a sampled trajectory \(\tau\), an auditor agent evaluates whether
each reference step is covered:
\[
A(\tau,d_i)\in\{0,1\}.
\]
The auditor sets \(A(\tau,d_i)=1\) only when the public interaction history
contains evidence that establishes the fact, anchor, or relation described by
\(d_i\). Tool observations provide the primary evidence. Assistant reasoning
is accepted only when it accurately cites or paraphrases evidence retrieved
earlier in the trajectory. Search queries, topical overlap, unsupported model
claims, and the final answer alone do not establish coverage, and each
reference step contributes at most once.

For reward computation, we separate steps that directly verify the gold answer
from those that establish intermediate anchors or relations:
\[
D_q^{\mathrm{answer}}
=
\{d_i\in D_q:
d_i \text{ directly verifies } y^\star\},
\qquad
D_q^{\mathrm{chain}}
=
D_q\setminus D_q^{\mathrm{answer}}.
\]

For process reward, let

\[
n_{\mathrm{chain}}(\tau)
=
\sum_{d_i\in D_q^{\mathrm{chain}}} A(\tau,d_i),
\qquad
n_{\mathrm{answer}}(\tau)
=
\sum_{d_i\in D_q^{\mathrm{answer}}} A(\tau,d_i)
\]
denote the numbers of covered intermediate and answer-supporting
reference steps, respectively. We define the process reward as
\[
R_{\mathrm{proc}}(\tau;D_q)
=
\min\left\{
\alpha n_{\mathrm{chain}}(\tau)
+
\beta\,\mathbb{I}[\operatorname{correct}(\tau)]
n_{\mathrm{answer}}(\tau),
C
\right\}.
\]

Intermediate coverage receives partial credit regardless of the final
outcome, whereas answer-supporting coverage contributes only when the
rollout is correct. The constant $C$ bounds the total process
contribution. Numerical settings are provided in Appendix \ref{app:training-details}.



The full reward used by PrimeSeeker-\(D_q\) is
\[
R(\tau;D_q)
=
R_{\mathrm{out}}(\tau)
+
R_{\mathrm{proc}}(\tau;D_q),
\]
where \(R_{\mathrm{out}}\) measures final-answer correctness.
PrimeSeeker-Outcome uses \(R_{\mathrm{out}}\) alone, isolating the effect of
reference-step coverage. Rewards are computed independently for each rollout
and normalized within each sampled group following standard GRPO. The auditor
prompt and remaining optimization details are provided in
Appendix~\ref{app:training-details}.
\section{Experiments}
\label{sec:experiments}

We evaluate PrimeSeeker through benchmark performance,
robustness under varying inference budgets, efficiency under a common
interaction budget, and the generated expert trajectories.

\subsection{Experimental Setup}
\label{sec:experimental-setup}

\paragraph{Benchmarks.}
We evaluate on BrowseComp-EN and BrowseComp-ZH
\citep{browsecomp,browsecomp_zh}, DeepSearch-2505 and DeepSearch-2510 from
xbench-DeepSearch~\citep{xbench_deepsearch}, and GAIA~\citep{gaia}.
BrowseComp evaluates target identification from sparse indirect clues,
xbench-DeepSearch emphasizes multi-step information seeking, and GAIA covers
broader retrieval and agentic reasoning. Due to evaluation cost, we use fixed
200-example subsets of BrowseComp-EN and BrowseComp-ZH for all PrimeSeeker
variants.

\paragraph{Baselines.}
Table~\ref{tab-main-public-results} includes advanced tool-augmented models
OpenAI o3, GPT-5, DeepSeek-V3.1, DeepSeek-V3.2, and GLM-4.7
\citep{openai_o3,gpt5,deepseek_v31,deepseek_v32,glm47}. We also compare with
selected open-source search agents of at least 30B parameters, including
Tongyi DeepResearch~\citep{tongyi_deepresearch},
DeepMiner~\citep{deepminer}, E-GRPO~\citep{egrpo},
WebSailor-V2~\citep{websailor_v2}, and OpenSeeker~\citep{openseeker,openseeker_v2}. MiroThinker is evaluated
separately under a shared interaction budget in
Section~\ref{sec:budgeted-reliability}.

\paragraph{Implementation details.}
All PrimeSeeker variants use Qwen3-30B-A3B-Thinking-2507~\citep{qwen3} and
the OpenSeeker live-web environment, with Serper~\citep{serper2025} for search
and Jina Reader~\citep{jina2025} for page retrieval.
\textbf{SFT.} We train for two epochs on 9,221 PrimeSeeker trajectories and 902
correct Chinese OpenSeeker trajectories. Tool observations are masked from the
loss, and all \(D_q\)-dependent guidance is removed before training.
\textbf{RL.} PrimeSeeker-Outcome and PrimeSeeker-\(D_q\) start from the same
SFT checkpoint and use the same questions and four rollouts per prompt, differing
only in whether reference-step coverage is included in the reward.
Each inference run is limited to 100 tool calls. Further details are provided in
Appendix~\ref{app:training-details}.

\paragraph{Evaluation metrics.}

For each question, we sample three independent runs. Pass@1 is the
average accuracy and Pass@3 is oracle
solution coverage. Answers enclosed by \texttt{<answer>} tags
are evaluated against the reference answer under an LLM-as-a-judge protocol.

\subsection{Main Benchmark Results}
\label{sec:main-results}

Table~\ref{tab-main-public-results} reports results on five benchmarks.
The first two blocks reproduce external reports under their respective
evaluation protocols. The final three rows report PrimeSeeker variants,
which achieve competitive performance among open-source search agents.
Relative to PrimeSeeker-SFT, PrimeSeeker-\(D_q\) improves Pass@1 on all
five benchmarks and Pass@3 on four, with Pass@3 unchanged on BS-ZH.
Compared with outcome-only RL, it improves both metrics on four
benchmarks but underperforms on BS-ZH.

\begin{figure*}[t]
    \centering
    \includegraphics[
        width=1.08\textwidth,
        keepaspectratio
    ]{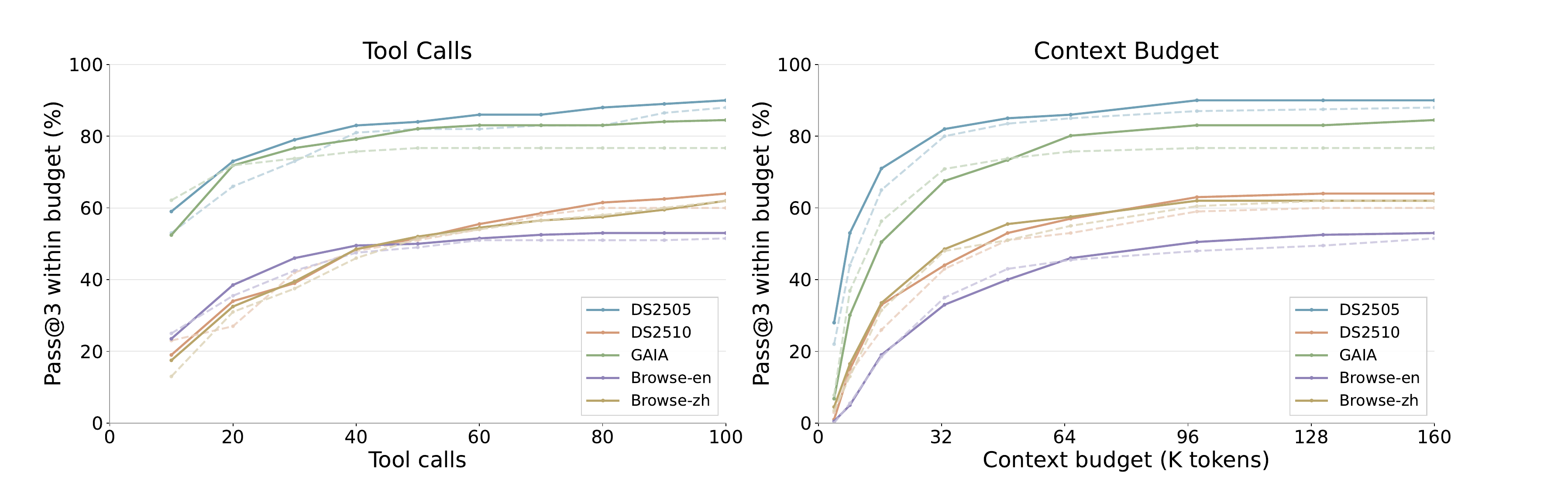}
    \caption{
    Pass@3 across tool call and context budgets. Solid lines denote
    PrimeSeeker-\(D_q\), and pale dashed lines denote PrimeSeeker-SFT. A
    question is solved if correct within the cutoff.
    }
    \label{fig:budget-sensitivity}
    \vspace{-0.4cm}
\end{figure*}

We retrospectively compute Pass@3 at tool-call and context-token cutoffs
from recorded runs of PrimeSeeker-SFT and PrimeSeeker-\(D_q\).
Figure~\ref{fig:budget-sensitivity} shows that PrimeSeeker-\(D_q\)
generally achieves higher solution coverage at matched cutoffs, with
most curves flattening around 60--80 calls and 64--96K tokens. 
These results show that its gains in solution coverage are already
evident below the maximum resource allowance.

\begin{table*}[t]
\caption{
Reported performance on five deep-search benchmarks. Underlined values are the best among reported models in each block; bold values are the best among
PrimeSeeker variants.
}
\label{tab-main-public-results}

\centering
\small
\renewcommand{\arraystretch}{1.12}
\setlength{\tabcolsep}{2.7pt}

\begin{tabularx}{\textwidth}{
@{}
>{\raggedright\arraybackslash}p{3.45cm}
| YY | YY | YY | YY | YY
@{}
}
\toprule

\raisebox{-0.9ex}{\bfseries Model}
& \multicolumn{2}{c|}{\bfseries BS-EN}
& \multicolumn{2}{c|}{\bfseries BS-ZH}
& \multicolumn{2}{c|}{\bfseries DS2505}
& \multicolumn{2}{c|}{\bfseries DS2510}
& \multicolumn{2}{c}{\bfseries GAIA}
\\[-0.2ex]

\cmidrule(lr){2-3}
\cmidrule(lr){4-5}
\cmidrule(lr){6-7}
\cmidrule(lr){8-9}
\cmidrule(lr){10-11}

& P@1 & P@3
& P@1 & P@3
& P@1 & P@3
& P@1 & P@3
& P@1 & P@3
\\
\midrule

\multicolumn{11}{l}{\itshape Advanced tool-augmented models} \\
\addlinespace[1.5pt]

OpenAI o3
    & 49.7 & --
    & 58.1 & --
    & 67.0 & --
    & --   & --
    & 70.5 & -- \\

DeepSeek-V3.1
    & 30.0 & --
    & 49.2 & --
    & 71.0 & --
    & -- & --
    & 63.1 & -- \\
    
GPT-5
    & 54.9 & --
    & 63.0 & --
    & 77.8 & --
    & \best{75.0} & --
    & \best{76.4} & -- \\

DeepSeek-V3.2
    & \best{67.6} & --
    & 65.0 & --
    & \best{78.0} & --
    & 55.7 & --
    & 75.1 & -- \\

GLM-4.7
    & 67.5 & --
    & \best{66.6} & --
    & 72.0 & --
    & 52.3 & --
    & 61.9   & -- \\

\addlinespace[2pt]
\midrule
\addlinespace[0.8pt]

\multicolumn{11}{l}{\itshape Open-source search agents (\(\geq 30\)B)} \\
\addlinespace[1.5pt]

Tongyi DeepResearch
    & 43.4 & \best{59.6}
    & 46.7 & \best{63.7}
    & 75.0 & \best{86.0}
    & \best{47.5} & --
    & 70.9 & \best{85.5} \\

DeepMiner-32B-RL
    & 33.5 & --
    & 40.0 & --
    & 62.0 & --
    & -- & --
    & 58.7 & -- \\

E-GRPO-30B
    & 12.9 & 21.0
    & 26.4 & 41.2
    & 46.7 & 66.0
    & -- & --
    & 48.5 & 65.1 \\

WebSailor-V2-30B
    & 35.3 & --
    & 44.1 & --
    & 73.7 & --
    & -- & --
    & \best{74.1} & -- \\

OpenSeeker
    & 29.5 & --
    & 48.4 & --
    & 74.0 & --
    & -- & --
    & -- & -- \\

OpenSeeker-V2
    & \best{46.0} & --
    & \best{58.1} & --
    & \best{78.0} & --
    & -- & --
    & -- & -- \\
    



\addlinespace[2.5pt]
\midrule

PrimeSeeker-SFT
    & 34.3 & 51.5
    & 40.8 & 62.0
    & 71.6 & 88.0
    & 40.3 & 60.0
    & 58.3 & 76.7 \\

+ Outcome Reward
    & 32.8 & 50.5
    & \textbf{47.5} & \textbf{63.5}
    & 74.0 & 84.0
    & 35.8 & 56.0
    & 56.3 & 69.9 \\

+ \(D_q\) Reward
    & \textbf{39.2} & \textbf{53.0}
    & 42.3 & 62.0
    & \textbf{76.0} & \textbf{90.0}
    & \textbf{41.0} & \textbf{64.0}
    & \textbf{62.1} & \textbf{84.5} \\

\bottomrule
\end{tabularx}
\end{table*}

\subsection{Search Efficiency under a Shared Interaction Budget}
\label{sec:budgeted-reliability}


We further compare PrimeSeeker with OpenSeeker-v1 and
MiroThinker~\citep{mirothinker}, rerunning all three systems on
100 questions from DS2505 and DS2510 under a shared cumulative budget
of 300 tool calls per question. All calls across context continuations
and up to five retries count toward this budget. Evaluation stops at
the first answer judged correct against the reference, or when the
budget or retry limit is reached. Budgeted Solve Rate measures the
fraction of questions solved. Consumption Avg.\ Calls averages cumulative
calls over all questions; Correct Avg.\ Calls averages calls in
successful attempts only, excluding earlier failures.


As shown in Table~\ref{tab-search-efficiency}, under the same interaction allowance, PrimeSeeker solves more questions on both datasets while consuming substantially fewer tool calls. This advantage is particularly pronounced on DS2510, where extended exploration becomes expensive. By focusing supervision on capability orientation rather than indirectly proxying difficulty through trajectory length, PrimeSeeker maximizes the utility of the fixed inference budget, achieving strong coverage without relying on prolonged interaction. This demonstrates that the strong deep search performance need not depend on exhaustive interaction.

\begin{table*}[t]
\caption{
Performance under a shared 300-call budget on DS2505 and
DS2510.
Pass@1 values for OpenSeeker-v1 and MiroThinker are from their original reports;
all remaining results are obtained by rerunning each
system with its original inference procedure.
}
\label{tab-search-efficiency}

\centering
\scriptsize
\renewcommand{\arraystretch}{1.12}
\setlength{\tabcolsep}{2.4pt}

\begin{tabularx}{\textwidth}{
@{}
>{\raggedright\arraybackslash}p{2.25cm}
| YYYY
| YYYY
@{}
}
\toprule

\raisebox{-1.45ex}[0pt][0pt]{\bfseries Model}
& \multicolumn{4}{c|}{\bfseries DS2505}
& \multicolumn{4}{c}{\bfseries DS2510}
\\[-0.2ex]

\cmidrule(lr){2-5}
\cmidrule(lr){6-9}

& \shortstack{Standard\\P@1}
& \shortstack{Budgeted\\Solve Rate}
& \shortstack{Consumption\\Avg.\ Calls}
& \shortstack{Correct\\Avg.\ Calls}
& \shortstack{Standard\\P@1}
& \shortstack{Budgeted\\Solve Rate}
& \shortstack{Consumption\\Avg.\ Calls}
& \shortstack{Correct\\Avg.\ Calls}
\\
\midrule

OpenSeeker-v1
& 74.0
& 83.0
& 34.5
& \textbf{16.2}
& --
& 57.0
& 109.5
& 46.4
\\

MiroThinker
& \textbf{77.2}
& 85.0
& 99.9
& 70.4
& \textbf{57.2}
& 58.0
& 144.2
& 72.6
\\

PrimeSeeker-\(D_q\)
& 76.0
& \textbf{91.0}
& \textbf{21.8}
& 16.3
& 41.0
& \textbf{68.0}
& \textbf{68.6}
& \textbf{33.5}
\\

\bottomrule
\end{tabularx}
\end{table*}

\subsection{Analysis of Constructed Expert Trajectories}
\label{sec:trajectory-quality}

Table~\ref{tab:trajectory-quality} compares corpus size, trajectory length,
tool usage, and exact within-trajectory repetition. PrimeSeeker has the
lowest average tool-call count and the lowest duplicate-query and
duplicate-visit rates among the compared corpora. Its average token length
remains above DeepResearch-9K but well below OpenSeeker and REDSearcher.
These statistics characterize the resulting demonstrations as relatively
compact in tool use and low in exact retrieval repetition. Full
distributional diagnostics are provided in
Appendix~\ref{app:trajectory-distributions}.

Complementing these corpus-level statistics, a controlled expert-generation analysis in Appendix~\ref{app:dq-guided-rollout} shows that reference-aligned $D_q$ highlights consistently improve accuracy across the evaluated teacher settings. Specifically, Qwen3 makes more calls, whereas DeepSeek-V4 makes fewer. This pattern is consistent with more effective evidence use rather than uniform search compression. Guidance may encourage further evidence gathering in shorter searches, while helping focus more extended exploration. In contrast, random highlights substantially degrade accuracy and increase tool use, showing that arbitrary observation-derived cues do not reproduce the gains.


\begin{table*}[]
\caption{Aggregate statistics of expert search trajectories.}
\vspace{-0.4cm}
\label{tab:trajectory-quality}
\begin{center}
\small
\setlength{\tabcolsep}{8pt}
\begin{tabular}{lrrrrr}
\toprule
\textbf{Dataset}
& \textbf{Samples}
& \textbf{Avg. Tools}
& \textbf{Avg. Tokens}
& \textbf{Search Dup.}
& \textbf{Visit Dup.}
\\
\midrule
DeepResearch-9K
&12,974
&25.0
&12.3k
&3.1\%
&13.1\%
\\

OpenSeeker
&11,677
&47.0
&73.9k
&1.6\%
&10.3\%
\\

REDSearcher
&10,001
&65.1
&56.4k
&0.2\%
&14.5\%
\\

PrimeSeeker
&9,221
&19.5
&37.8k
&0.1\%
&8.0\%
\\
\bottomrule
\end{tabular}
\end{center}
\vspace{-0.4cm}
\end{table*}

\section{Related Work}
\label{sec:related-work}

\paragraph{Deep-search agents.}
Retrieval-augmented generation established external evidence as a complement to
parametric knowledge, while later methods made retrieval adaptive to the
model's evolving reasoning state
\citep{NEURIPS2020_6b493230,trivedi-etal-2023-interleaving,
asai2024selfrag,NEURIPS2025_566aad4f}. ReAct formulated tool use as an iterative
reasoning--action--observation process and became the basis of many web-search
agents~\citep{yao2023react}. Search-R1 and R1-Searcher demonstrated that this
behavior can be acquired through outcome-supervised reinforcement learning,
while recent DeepResearch systems combine supervised trajectories, policy
optimization, and extended inference horizons
\citep{jin2025searchr,song2025r1searcherincentivizingsearchcapability,
tongyi_deepresearch,mirothinker,redsearcher}.
These works mainly improve agent architectures, post-training, and
inference-time search. PrimeSeeker instead studies the supervision encoded by
constructed search problems, focusing on how an indirectly specified target is
recovered as a retrieval anchor and enables the next information demand.

\paragraph{Synthetic search-task construction.}
Classical multi-hop QA constructs questions around linked supporting facts or
composed single-hop questions
\citep{yang-etal-2018-hotpotqa,trivedi-etal-2022-musique}. Recent search-agent
pipelines create harder, implicit information needs through web-grounded graph
expansion, relational subgraph sampling, entity obfuscation, and dispersed
evidence
\citep{li2025websailornavigatingsuperhumanreasoning,websailor_v2,
openseeker,mirothinker,redsearcher}. WebSailor-V2 broadens structural and
uncertainty coverage through densely connected graphs, topology-aware sampling,
and information obfuscation~\citep{websailor_v2}. OpenSeeker and related
easy-to-hard pipelines iteratively expand seed questions or local web structures
into more challenging search problems~\citep{openseeker}.

These approaches substantially improve the scale, diversity, and difficulty of
synthetic search data. However, graph size, hop count, and realized trajectory
length do not uniquely determine the retrieval capability exercised by a
question. Selective clues, co-located evidence, or parametric knowledge may
collapse an apparently complex problem, motivating shortcut detection and
empirical difficulty filtering
\citep{fort_searcher,song2026demystifying}. PrimeSeeker adopts a complementary
granularity by organizing construction around the local unit
\(C(a_i)\Rightarrow a_i\xrightarrow{r_i}z_i\). This unit couples the recovery
of a descriptively specified target with the relation demand enabled by that
target, while allowing different agents to realize the capability through
different search paths.

\paragraph{Expert supervision and process rewards.}
Search-agent SFT commonly relies on successful trajectories sampled from
stronger teachers
\citep{NEURIPS2025_566aad4f,
li2025websailornavigatingsuperhumanreasoning,redsearcher}. Subsequent methods
improve these demonstrations during generation or through post-hoc processing.
OpenSeeker summarizes earlier observations to assist teacher generation and
restores the original observations before student training, whereas WebClipper
prunes redundant branches and rewrites the retained reasoning
\citep{openseeker,wang-etal-2026-webclipper}. Verification-based systems also
use external feedback to revise actions or final answers
\citep{mirothinker,zhu2026marcodeepresearchunlockingefficient}.

A related line of work reuses information from task construction during policy
optimization. E-GRPO rewards matches to retained ground-truth entities, while
Search-P1 rewards consistency with reference plans derived from successful
solutions
\citep{egrpo,xia-etal-2026-search}. PrimeSeeker instead derives the reference evidence skeleton \(D_q\) jointly with each question
and uses it for expert guidance and reference rewards, with all auxiliary
annotations removed before SFT.

\section{Conclusion} 
\label{sec:conclusion} 

We introduced latent anchor reasoning as a capability-oriented formulation of
deep search, decomposing local retrieval into latent-anchor resolution and
relation transfer. PrimeSeeker operationalizes this formulation through
web-grounded anchor structures from which questions and reference evidence
skeletons \(D_q\) are jointly derived. During expert generation, \(D_q\)
provides strictly extractive, observation-grounded highlights without exposing
unseen evidence or prescribing a canonical search path. These annotations are
removed before SFT, while the same construction-time structure is later reused
to audit reference-step coverage during reinforcement learning. PrimeSeeker
therefore connects task construction, expert synthesis, and policy optimization
through a shared representation of the retrieval capabilities required by each
question.

Using 9,221 generated trajectories, PrimeSeeker achieves strong performance across five deep-search benchmarks. Its demonstrations exhibit low retrieval redundancy, reference-aligned guidance improves expert generation while random highlights substantially degrade it, and fixed-budget evaluation shows that strong solution coverage need not rely on extended search trajectories. These results support explicit local retrieval capabilities and reusable construction-time structure as a more direct basis for search-agent supervision.

\newpage
\bibliographystyle{plainnat}
\bibliography{references}

\appendix
\newpage

\section{Appendix Overview}
\label{app:overview}

This appendix provides implementation details, controlled analyses,
qualitative examples, and reproducibility information omitted from the
main paper. Section~\ref{app:training-details} details the training, inference,
tool environment, and auxiliary-model configurations.
Section~\ref{app:dq-guided-rollout} presents a controlled study of unguided,
random-highlight, and $D_q$-guided expert generation.
Section~\ref{app:trajectory-distributions} analyzes the trajectory-length,
tool-use, and retrieval-redundancy distributions of the constructed expert
trajectories. Section~\ref{app:qa-cases} provides case studies illustrating
how web-grounded anchor structures give rise to aligned questions $q$ and
reference evidence skeletons $D_q$. Section~\ref{app:prompt-templates}
provides the prompt templates for observation-grounded evidence extraction
and reference-step coverage auditing. Section~\ref{app:test-rollout-cases}
presents a representative test-time rollout from the trained PrimeSeeker
policy. Section~\ref{app:release} summarizes the released artifacts and
discusses limitations.
\section{Training and Deployment Details}
\label{app:training-details}

\paragraph{Supervised fine-tuning.}
We initialize PrimeSeeker from
\texttt{Qwen3-30B-A3B-Thinking-2507}. The training corpus contains 9,221
PrimeSeeker expert trajectories and 902 correct Chinese trajectories from
OpenSeeker~\citep{openseeker}, yielding 9,589 training instances and 534
validation instances. Tool observations are retained as context but masked from
the language-modeling loss. We train for two epochs with a global batch size of
64, a maximum sequence length of 196,608 tokens, and a peak learning rate of
\(1\times10^{-5}\) followed by cosine decay. The resulting checkpoint
initializes both reinforcement-learning variants.

\paragraph{Reinforcement learning.}
PrimeSeeker-Outcome and PrimeSeeker-\(D_q\) are initialized from the same
SFT checkpoint and trained on the same set of question--answer pairs
generated using the QA construction procedure described in Section~\ref{sec:qa-construction}.
Both variants are trained with GRPO for 64 steps. Each step uses a batch
of 16 questions, with four rollouts sampled per question to form a GRPO
group, yielding 64 rollouts per step. We use an actor learning rate of
\(2\times10^{-6}\), a PPO clipping ratio of \(0.2\), and a KL coefficient
of \(0.001\). PrimeSeeker-Outcome uses final-answer correctness alone,
whereas PrimeSeeker-\(D_q\) additionally uses the reference-step coverage
reward defined in Section~\ref{sec:training}. Invalid formats, repeated
generations, context failures, and tool-budget violations receive a
failure reward.
For all PrimeSeeker-$D_q$ experiments, we set
$\alpha=0.10$, $\beta=0.20$, and $C=0.50$.

\paragraph{Rollout and inference.}
RL rollouts are sampled with temperature \(1.0\) and top-\(p=0.95\).
For benchmark evaluation, we use temperature \(0.8\), top-\(p=0.95\), and a
repetition penalty of \(1.05\). Unless otherwise specified, each trajectory is
limited to 100 tool calls.

\paragraph{Tool environment and auxiliary models.}
Following OpenSeeker~\citep{openseeker}, all experiments use the same live-web
interaction environment and evaluation protocol, with
Serper~\citep{serper2025} for web search and
Jina Reader~\citep{jina2025} for page fetching. The construction operators,
closed-book filter, \(\operatorname{Extract}_{D}\),
\(\operatorname{Extract}_{k}\), trajectory verifier, reference-step auditor,
and answer evaluator are served by the same
\texttt{Qwen3-30B-A3B-Instruct-2507} deployment with a 230K-token context
window.

\section{Effect of Observation-grounded
\texorpdfstring{\(D_q\)}{Dq} Guidance}
\label{app:dq-guided-rollout}

We compare unguided, \(D_q\)-guided, and random-highlight generation using
Qwen3 and DeepSeek-V4 on questions with \(K\in\{2,3\}\) independently sourced
specifications per anchor. The random control preserves the key-field format
but samples two to five content words from the current observation independently
of \(q\) and \(D_q\); all other settings are identical. As shown in
Table~\ref{tab:dq-guided-rollout}, \(D_q\)-guided highlights improve accuracy
by 2\%--7\%, whereas random highlights reduce it by 23\%--35\% and add
25--65 tool calls. This contrast shows that the benefit arises from reference
alignment rather than the additional key field itself.
\begin{table*}[t]
\caption{
Comparison of random and \(D_q\)-guided highlights against unguided expert generation.
}
\label{tab:dq-guided-rollout}

\centering
\small
\renewcommand{\arraystretch}{1.10}
\setlength{\tabcolsep}{2.8pt}

\begin{tabularx}{\textwidth}{
@{}
>{\raggedright\arraybackslash}p{2.45cm}
c
|
YY
|
YYYY
|
YYYY
@{}
}
\toprule

\raisebox{-1.0ex}{\bfseries Expert}
& \raisebox{-1.0ex}{\bfseries \(K\)}
& \multicolumn{2}{c|}{\bfseries Unguided}
& \multicolumn{4}{c|}{\bfseries Random Highlight}
& \multicolumn{4}{c}{\bfseries \(D_q\)-guided}
\\[-0.2ex]

\cmidrule(lr){3-4}
\cmidrule(lr){5-8}
\cmidrule(lr){9-12}

& & Acc. & Calls
& Acc. & \(\Delta\)Acc. & Calls & \(\Delta\)Calls
& Acc. & \(\Delta\)Acc. & Calls & \(\Delta\)Calls
\\
\midrule

Qwen3
& 2
& 52.0 & 6.45
& 28.0 & -24.0 & 70.52 & +64.07
& 54.0 & +2.0 & 9.09 & +2.64
\\

Qwen3
& 3
& 57.0 & 7.01
& 34.0 & -23.0 & 72.10 & +65.09
& 61.0 & +4.0 & 9.87 & +2.86
\\

\addlinespace[1.5pt]
\midrule
\addlinespace[1.5pt]

DeepSeek-V4
& 2
& 51.0 & 24.67
& 24.0 & -27.0 & 50.12 & +25.45
& 56.0 & +5.0 & 21.99 & -2.68
\\

DeepSeek-V4
& 3
& 59.0 & 28.68
& 24.0 & -35.0 & 57.18 & +28.50
& 66.0 & +7.0 & 24.59 & -4.09
\\

\bottomrule
\end{tabularx}
\end{table*}

\section{Trajectory Distribution Diagnostics}
\label{app:trajectory-distributions}

\begin{figure*}[t]
\centering
\begin{subfigure}[t]{0.48\linewidth} 
    \centering
    \includegraphics[width=\linewidth]{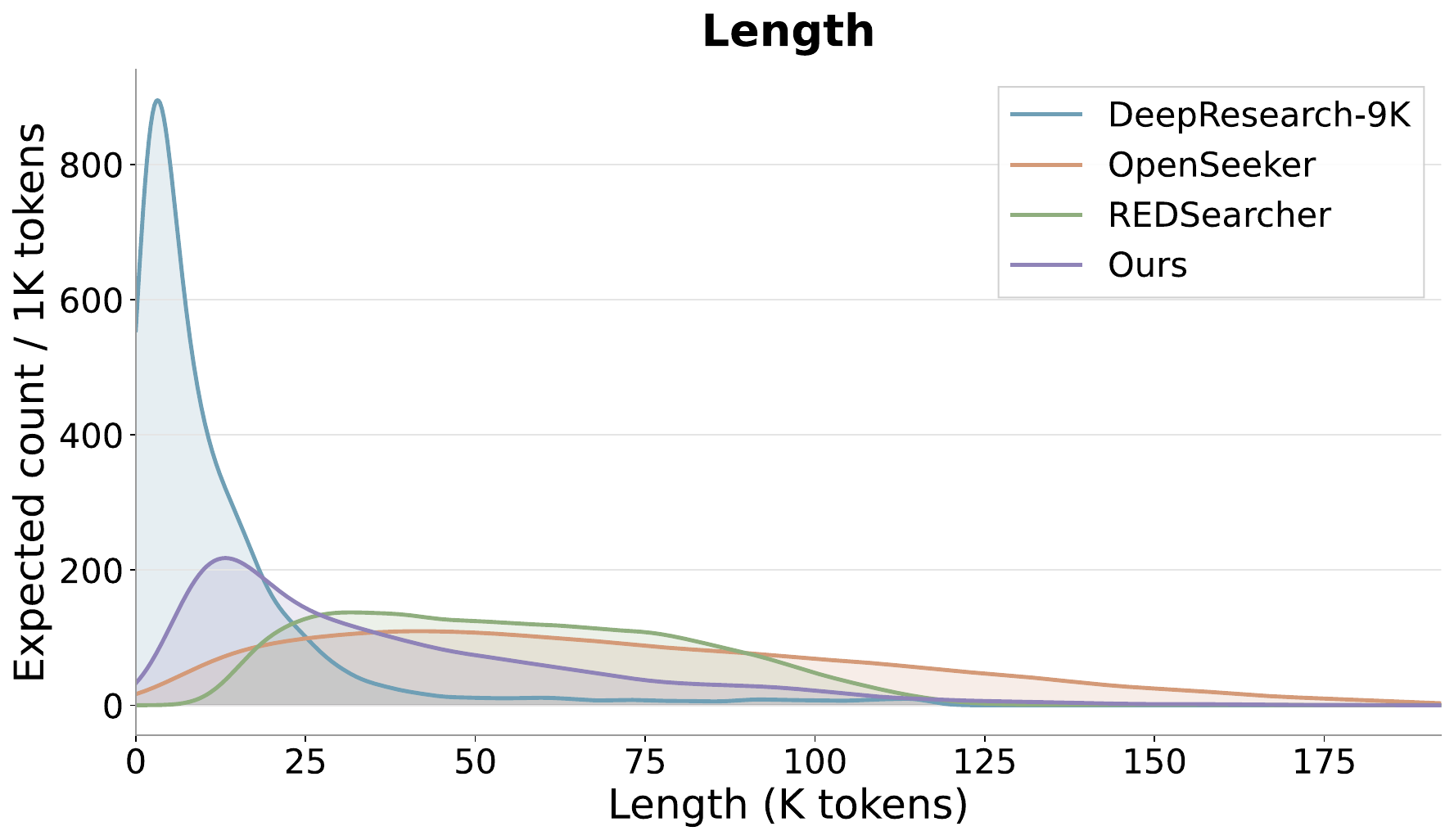}
    \caption{Trajectory Length}
    \label{fig:len_dist}
\end{subfigure}
\hfill 
\begin{subfigure}[t]{0.48\linewidth}
    \centering
    \includegraphics[width=\linewidth]{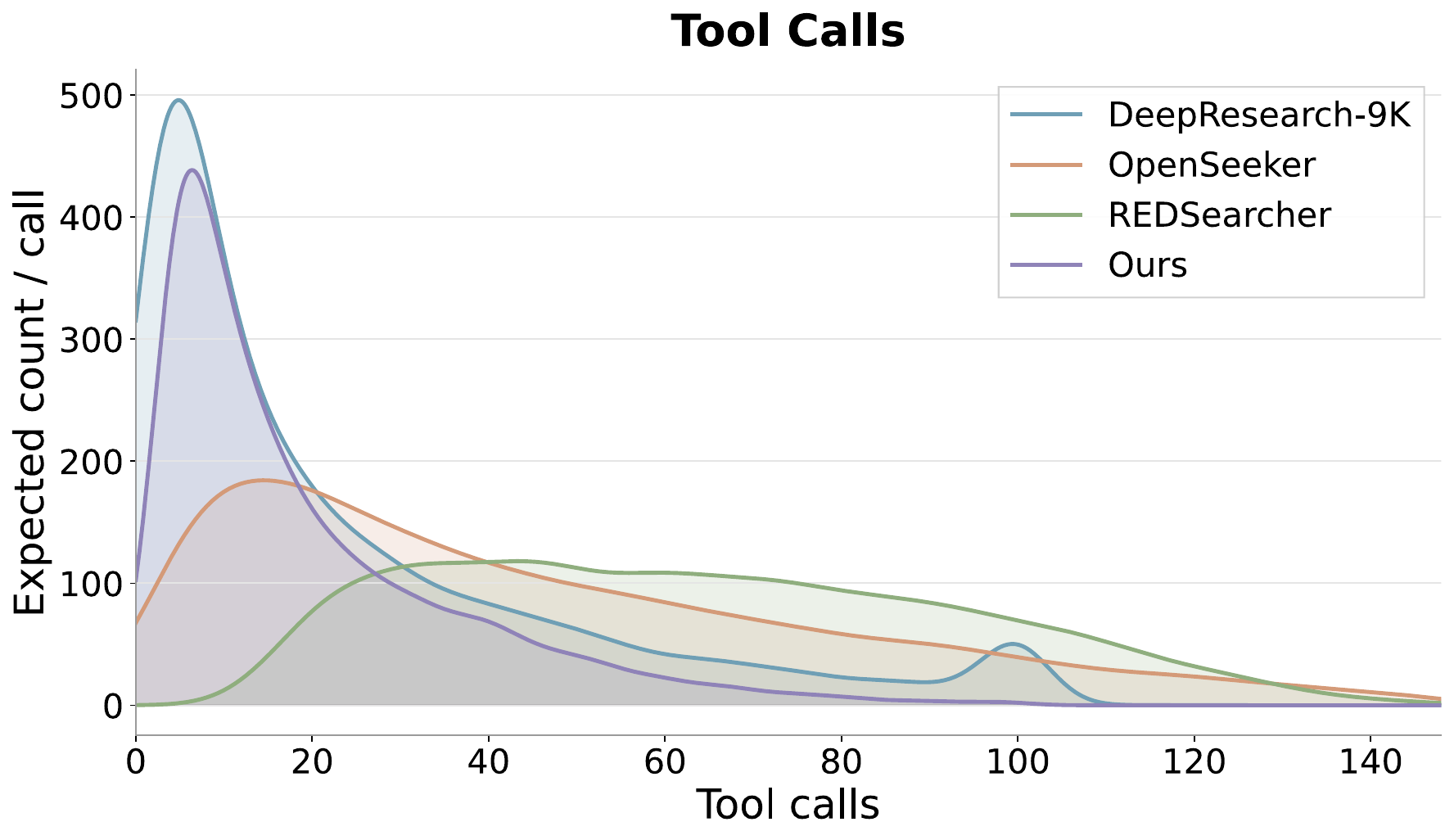}
    \caption{Tool Calls}
    \label{fig:tool_calls}
\end{subfigure}

\vspace{1em} 
\begin{subfigure}[t]{0.48\linewidth}
    \centering
    \includegraphics[width=\linewidth]{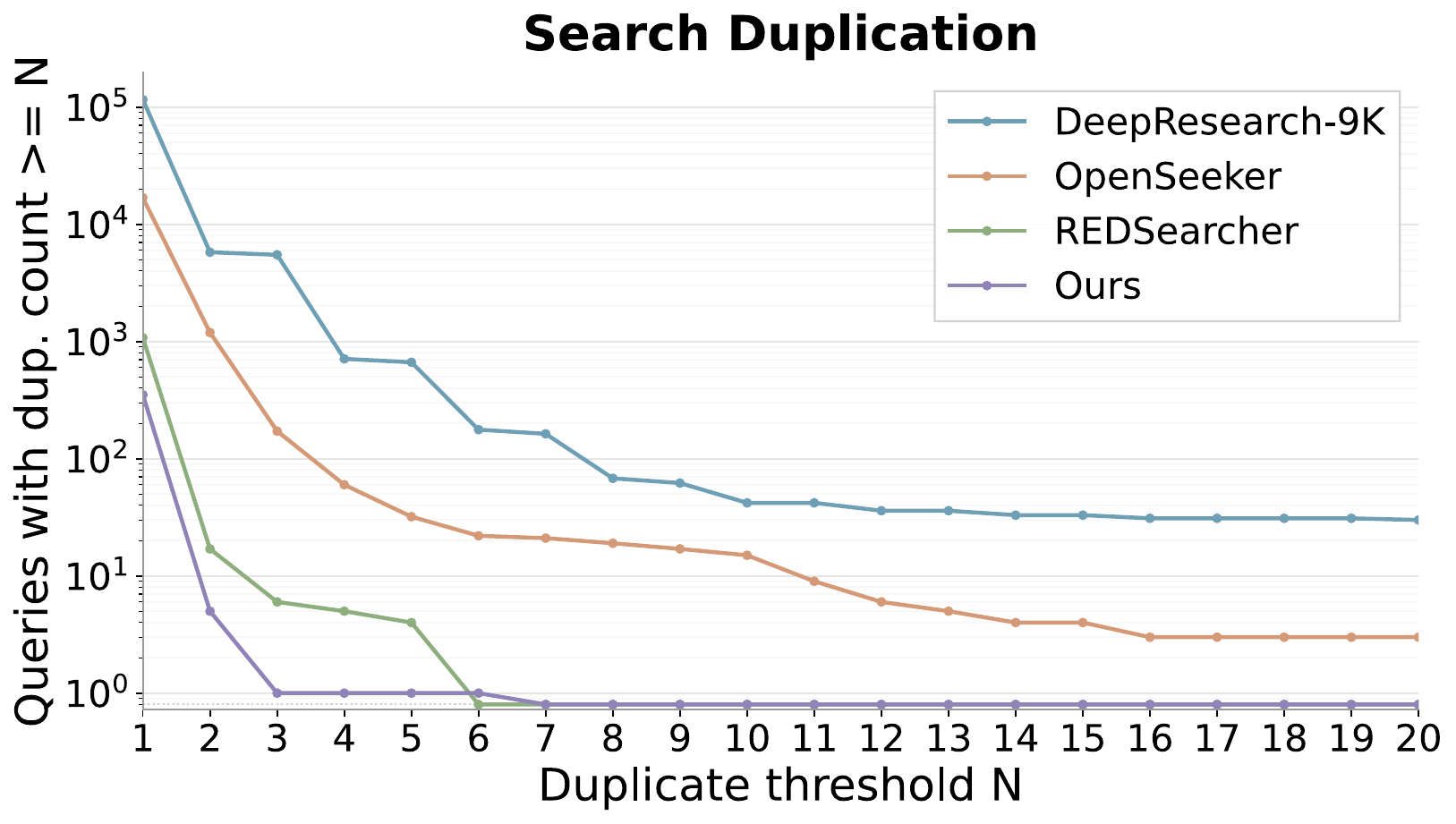}
    \caption{Duplicate Searches}
    \label{fig:search_dup}
\end{subfigure}
\hfill
\begin{subfigure}[t]{0.48\linewidth}
    \centering
    \includegraphics[width=\linewidth]{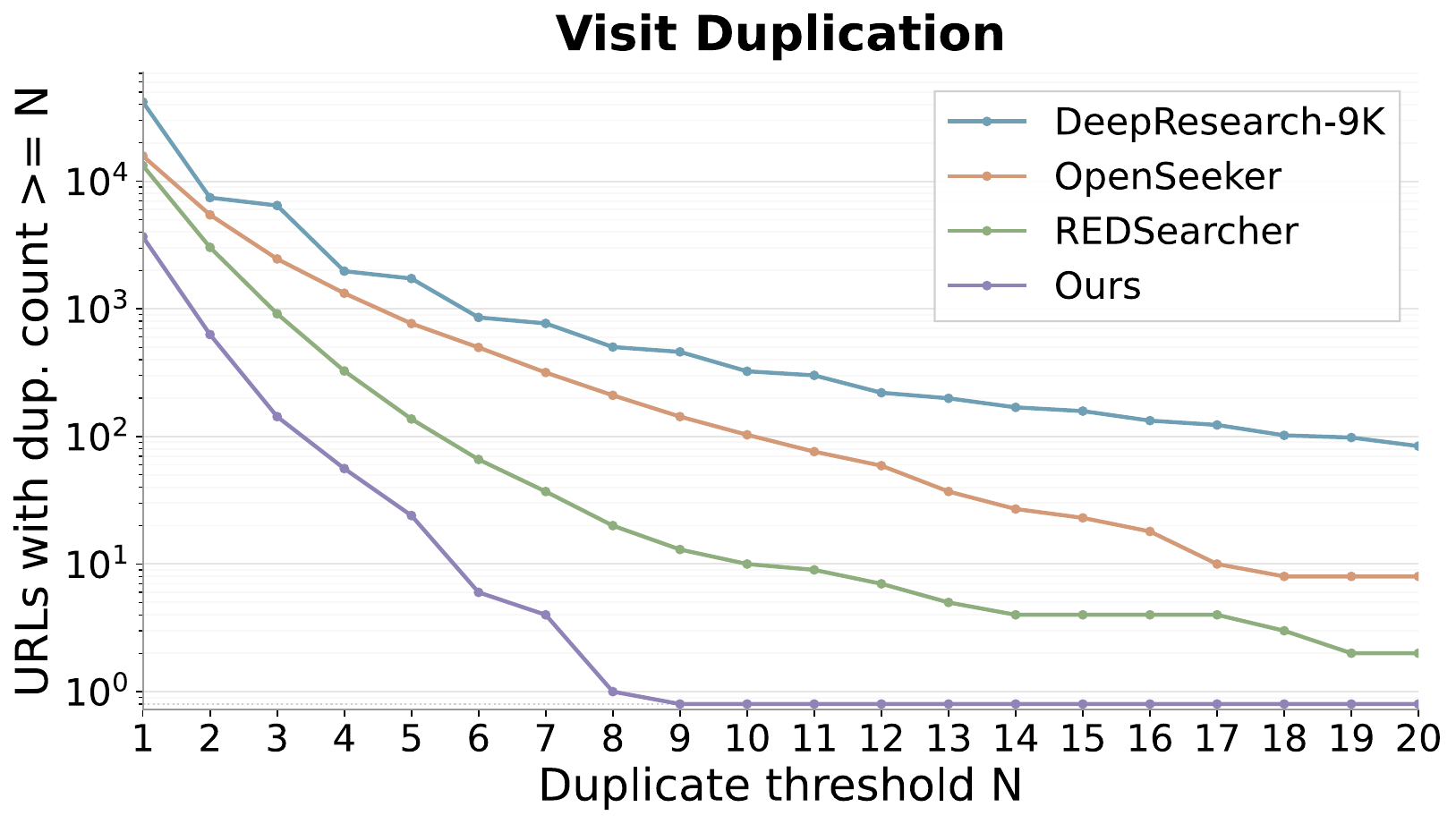}
    \caption{Duplicate URLs}
    \label{fig:url_dup}
\end{subfigure}

\caption{
Distributional diagnostics of expert trajectories.
(a) Trajectory length distribution, shown as KDE-smoothed expected sample count
per 1K-token window.
(b) Tool-call distribution, shown as KDE-smoothed expected sample count per
one-tool-call window.
(c) CCDF of duplicated search-query counts within trajectories.
(d) CCDF of duplicated visited-URL counts under the same criterion.
PrimeSeeker trajectories occupy moderate tool-use ranges and exhibit
lighter duplicate-retrieval tails, indicating more compact and less redundant
expert supervision.
}
\label{fig:trajectory-distribution}
\end{figure*}

\paragraph{Distributional analysis.}
To complement the aggregate statistics,
Figure~\ref{fig:trajectory-distribution} visualizes the complete trajectory
distributions. Figures~\ref{fig:len_dist} and~\ref{fig:tool_calls} show
KDE-smoothed expected sample counts over trajectory length and tool-call count,
respectively. For trajectory length, the vertical axis represents the expected
number of samples within a nearby 1K-token window after smoothing. For tool
calls, it represents the expected number of trajectories around each tool-call
count.

DeepResearch-9K is concentrated in very short trajectories, whereas OpenSeeker
has the heaviest long-context tail. REDSearcher is concentrated in the
mid-to-long range. PrimeSeeker occupies the short-to-middle range and
has a substantially lighter long tail than OpenSeeker and REDSearcher. The
tool-call distribution exhibits a similar pattern: DeepResearch-9K and
PrimeSeeker concentrate on lower tool-call ranges, OpenSeeker has a
broader middle-range distribution, and REDSearcher is shifted toward heavier
tool usage. The small concentration of DeepResearch-9K trajectories near its
maximum tool-call region may reflect a tool-budget boundary.

Figures~\ref{fig:search_dup} and~\ref{fig:url_dup} show complementary cumulative
distributions of duplicate searches and duplicate page visits. The horizontal
axis gives the duplicate count within a trajectory, while the vertical axis
reports the number of trajectories whose duplicate count is at least that
value. Lower curves therefore indicate lighter redundancy tails.
PrimeSeeker exhibits consistently lighter duplicate-search and
duplicate-visit tails, providing distribution-level evidence that its compactness
is associated with less low-information retrieval rather than premature
termination alone.

Together, these statistics show that PrimeSeeker does not rely on globally
inflated trajectories and that its demonstrations contain substantially less
repeated retrieval. Their downstream utility is evaluated through the trained
policy results in Section~\ref{sec:main-results}.

\section{QA Construction Case Study}
\label{app:qa-cases}

We present two construction examples showing the aligned outputs \(q\) and
\(D_q\) derived from a shared web-grounded anchor structure. Blue text denotes
descriptive specifications, orange text denotes latent or recovered anchors,
and green text denotes the target answer. Each specification is extracted from
an independently retrieved source page; the complete provenance records are
included in the released data.

\begin{CaseStudyBox}{Case 1: Isle of Skye}

\begin{QuestionBlock}
\textbf{Question \(q\).}
In a 17th-century play, a character forces two others to perform a temporary
marriage ceremony. This ceremony has a historical counterpart in a European
country, where on a specific island, such a ceremony was recorded in the late
1600s. What is the name of that island?
\end{QuestionBlock}

\CaseLabel{Reference evidence skeleton \(D_q\)}
\begin{ReferenceSteps}
  \RefStep{\textsc{Chain} \(d_1\)}
  {\textit{The Witch} is the 17th-century play described in the question.}

  \RefStep{\textsc{Chain} \(d_2\)}
  {Antonio forces Francisca and Aberzanes to perform a handfast ceremony.}

  \RefStep{\textsc{Chain} \(d_3\)}
  {Handfasting was associated with temporary or probationary marriage in
  later Scottish tradition.}

  \RefStep{\textsc{Answer} \(d_4\)}
  {A late-17th-century account records this custom on the Isle of Skye.}
\end{ReferenceSteps}

\CaseLabel{Latent-anchor sequence}
\begin{sloppypar}
\ChainClue{17th-century play}
\ChainArrow
\ChainAnchor{\textit{The Witch}}
\ChainArrow
\ChainClue{temporary marriage ceremony}
\ChainArrow
\ChainAnchor{handfasting}
\ChainArrow
\ChainAnchor{Scottish tradition}
\ChainArrow
\ChainAnswer{Isle of Skye}
\end{sloppypar}
\end{CaseStudyBox}

\begin{CaseStudyBox}{Case 2: Castruccio Castracani}

\begin{QuestionBlock}
\textbf{Question \(q\).}
A 13th-century treatise on human equality is linked to a historic mass
emancipation of serfs in a major Italian city-state. Decades later, that city
supported Florence against a Ghibelline ruler from a noble family of Lucca.
This ruler had been a leading figure in the 1315 victory at Montecatini. Who
was he?
\end{QuestionBlock}

\CaseLabel{Reference evidence skeleton \(D_q\)}
\begin{ReferenceSteps}
  \RefStep{\textsc{Chain} \(d_1\)}
  {\textit{Liber Paradisus} is the 13th-century treatise associated with human
  equality.}

  \RefStep{\textsc{Chain} \(d_2\)}
  {The document records the collective emancipation of serfs by Bologna.}

  \RefStep{\textsc{Chain} \(d_3\)}
  {The city-state referred to in the question is Bologna.}

  \RefStep{\textsc{Chain} \(d_4\)}
  {Bologna supported Florence against a Ghibelline ruler from the
  Antelminelli/Castracani family of Lucca.}

  \RefStep{\textsc{Answer} \(d_5\)}
  {Castruccio Castracani was a leading figure in the 1315 Ghibelline victory
  at Montecatini.}
\end{ReferenceSteps}

\CaseLabel{Latent-anchor sequence}
\begin{sloppypar}
\ChainClue{13th-century treatise on human equality}
\ChainArrow
\ChainAnchor{\textit{Liber Paradisus}}
\ChainArrow
\ChainAnchor{Bolognese mass emancipation}
\ChainArrow
\ChainAnchor{Bologna}
\ChainArrow
\ChainAnchor{Ghibelline ruler from Lucca}
\ChainArrow
\ChainAnswer{Castruccio Castracani}
\end{sloppypar}

\end{CaseStudyBox}

\section{Prompt Templates}
\label{app:prompt-templates}

We provide the prompts for the operators that instantiate and use the reference
evidence skeleton. All operators use the auxiliary model specified in
Section~\ref{app:training-details}. The exact upstream prompts used to expand
anchor structures and verbalize questions are not included. Instead, we specify
their input--output contracts and release the resulting questions, reference
evidence skeletons, trajectories, and source provenance.

\subsection{Key Observation Extraction}
\label{app:key-observation-prompt}

\paragraph{External validation of extracted highlights.}
The \texttt{key\_observation} entries returned by the prompt below are
treated as candidate highlights. Before they are used for trajectory
generation, the surrounding pipeline checks each candidate by direct
string matching against the current raw observation. Candidates not
found verbatim are discarded, as are candidates containing a direct
string match to the gold answer. Only the surviving spans are used as
auxiliary highlights; if none survive, the original observation is
passed to the expert without additional guidance. This post-processing
filters the highlights without modifying the retrieved observation.
\begin{PromptBox}{Prompt: Key Observation Extraction}
[SYSTEM]

## Role

You are an information filter for tool-augmented question answering.
The reference steps describe the evidence needed to solve the question.
Your task is to extract matching evidence from the current tool observation
and conservatively identify repeated search keywords or visit goals.

## Procedure

### 1. Align the observation with the reference steps

- The raw observation must explicitly support the same specific fact described
  by a reference step, such as an entity, name, date, event, or relation.
  Broad topical similarity is insufficient.
- If no reference step is matched, set `matched_reference_index` and
  `matched_reference_step` to null.

### 2. Extract the key observation

- Copy every sentence from the raw observation that provides direct evidence
  for the matched reference step.
- Copy the evidence verbatim. Do not infer, summarize, rewrite, or repeat it.
- `matched_reference_step` must be copied verbatim from the reference steps.
- `key_observation` must contain verbatim sentences from the raw observation.
- The matched reference step and key observation must come from different
  sources and must not be identical.
- If the evidence is uncertain, unsupported, or not a verbatim substring of
the raw observation, set `key_observation` to null.


## Hard Rules

- Return strict JSON only.
- Do not include markdown, explanations, or reasoning outside the JSON.
- Use null for any unavailable field.

## Output Format

Return JSON with exactly the following keys:

{
  "matched_reference_index": 0,
  "matched_reference_step": "<verbatim reference step or null>",
  "key_observation": ["<verbatim sentence from raw_observation>"],
  "matched_reason": "<brief explanation of the evidence match or null>"
}

[USER]

Question:
{question}

Reference steps:
{reference_steps}

Current search keywords or visit goal:
{search_keywords_or_visit_goal}

Raw observation:
{raw_observation}

Previous tool keywords:
{history_tool_keywords}
\end{PromptBox}

\subsection{Reference-Step Coverage Auditing}
\label{app:dq-matching-prompt}

\begin{PromptBox}{Prompt: Reference-Step Coverage Auditing}
[SYSTEM]

## Role

You are an evidence auditor for a tool-using search agent. You may inspect
hidden reference milestones, but you must evaluate coverage using only evidence
available in the public trajectory. Hidden anchors and unretrieved facts must
never be attributed to the agent.

## Evidence Types

- `descriptor`: evidence supporting a descriptive clue for a hidden anchor
- `anchor`: evidence identifying or substantially narrowing a hidden anchor
- `transition`: evidence supporting a verifiable relation between anchors
- `answer`: evidence directly identifying or verifying the final answer

## Evaluation Procedure

### 1. Audit each reference step

For every numbered reference step, determine:

- its evidence type;
- whether it is covered;
- the public trajectory event that supports it;
- the minimal supporting quote or paraphrase; and
- whether the evidence contributes to the reasoning chain or directly supports
  the answer.

### 2. Collect effective evidence

Record each effective evidence item separately and identify:

- the trajectory step where it appears;
- the reference steps it supports;
- its evidence role and evidence type;
- its importance for the search decision; and
- the next action it enables.

Use the following decision levels:

- `weak`: topical evidence that does not materially reduce uncertainty
- `useful`: evidence that enables a better query or anchor transition
- `decisive`: evidence that identifies a required anchor or directly verifies
  the answer

Use the following action implications:

- `continue_search`
- `transition_query`
- `answer_ready`
- `no_progress`

Set `answer_ready` only when the public evidence is sufficient to answer the
question.

### 3. Assess trajectory reliability

Indicate whether the audit relies on an unsupported model guess, whether the
trajectory contains repeated or irrelevant behavior, and the overall confidence
of the audit.

## Hard Rules

- Evaluate only the public trajectory.
- Return compact, strict JSON only.
- Do not include markdown, explanations, or reasoning outside the JSON.
- Include one item in `reference_step_audit` for every numbered reference step.

## Output Format

Return JSON with exactly the following structure:

{
  "reference_step_audit": [
    {
      "ref_step": 1,
      "milestone_type": "descriptor | anchor | transition | answer | unknown",
      "covered": true,
      "event_type": "tool_response | model_message | null",
      "public_evidence": "<minimal quote or paraphrase from the public trajectory or null>",
      "evidence_role": "chain | answer | none"
    }
  ],
  "all_effective_evidence": [
    {
      "step": 1,
      "event_type": "tool_response | model_message",
      "ref_steps": [1],
      "evidence": "<minimal public evidence>",
      "evidence_role": "chain | answer | none",
      "milestone_type": "descriptor | anchor | transition | answer | unknown",
      "decision_level": "weak | useful | decisive",
      "next_action_implication": "continue_search | transition_query | answer_ready | no_progress",
      "why_effective": "<brief justification>"
    }
  ],
  "used_model_guess_as_evidence": false,
  "off_chain_or_redundant": false,
  "confidence": 0.0
}

[USER]

Question:
{question}

Ground-truth answer:
{ground_truth}

Reference steps:
{reference_step}

Predicted final answer:
{predicted_answer}

Audit policy:
{audit_policy}

Step-labeled public trajectory:
{trajectory}
\end{PromptBox}


\section{Test-Time Rollout Case Study}
\label{app:test-rollout-cases}

We present a representative test-time rollout produced by the trained PrimeSeeker
policy preserving the original order. The conversations are reproduced in their original order without
rewriting or reordering intermediate reasoning, tool calls, and
observations.

\begin{RolloutCaseHeader}
{Case: Identifying a television episode from indirect evidence}

\textbf{Dataset:} BrowseComp-EN
\hfill
\textbf{Model:} PrimeSeeker-\(D_q\)

\textbf{Tool calls:} 6
\hfill
\textbf{Result:} Correct

\textbf{Reference answer:}
\emph{3rd Rock from the Sun; Season 1; Episode 2; Post-Nasal Dick}
\end{RolloutCaseHeader}

\begin{QuestionTurn}
question:"Provide the name of the TV series along with the season, episode number, and episode title that meet the following criteria:  The episode aired in January between 1994 and 1997, inclusive.  One of the genres of the TV series is comedy.  In the episode, one character accompanies another, who is considered their love interest in the series, to a wedding.  A few characters fall sick for the first time and become convinced they are dying.  One filming mistake in the episode involves a character placing their feet in a pot filled with liquid, but by the end of the scene, their socks appear dry."

\end{QuestionTurn}

\begin{AssistantTurn}
assistant:"
<think>
We need to identify a TV series episode that meets several specific criteria.
...
</think>
<tool_call>
[
  {
    "function": {
      "name": "search",
      "arguments": {
        "query": [
          "TV episode aired January 1994 to 1997 comedy wedding"
        ]
      }
    }
  }
]
</tool_call>"
\end{AssistantTurn}

\begin{UserTurn}
user:"Observation from search:
### A Google search for 'TV episode aired January 1994 to 1997 comedy wedding' found 10 results:
..."
\end{UserTurn}

\begin{AssistantTurn}
assistant:"
<think>
The search results did not directly identify the episode. Let me refine the
query using the filming mistake.
</think>
<tool_call>
[
  {
    "function": {
      "name": "search",
      "arguments": {
        "query": [
          "\"feet in a pot\" \"dry socks\" TV episode mistake"
        ]
      }
    }
  }
]
</tool_call>"
\end{AssistantTurn}

\begin{UserTurn}
user:"Observation from search:
### A Google search for '\"feet in a pot\" \"dry socks\" TV episode mistake' found 0 results:

"
\end{UserTurn}

\begin{AssistantTurn}
assistant:"
<think>
No direct results. Let me try searching for the filming mistake in a different way, perhaps by looking for continuity errors in television episodes.
</think>
<tool_call>
[
  {
    "function": {
      "name": "search",
      "arguments": {
        "query": [
          "continuity error TV episode feet in pot dry socks"
        ]
      }
    }
  }
]
</tool_call>"
\end{AssistantTurn}

\begin{UserTurn}
user:"Observation from search:
### A Google search for 'continuity error TV episode feet in pot dry socks' found 10 results:

1. [3rd Rock from the Sun (1996) revealing mistakes in Post Nasal Dick](https://www.moviemistakes.com/tv4916/episode23240/revealing)

Harry put his feet in a pot of warm chicken soup, yet at the end of that scene when they hug each other Harry lifts his feet up and you can see his socks are ...

2. ..."
\end{UserTurn}

\begin{AssistantTurn}
assistant:"
<think>
I found a likely match from 3rd Rock from the Sun. The filming mistake matches
the clue, but I still need to verify the episode title, season, episode number,
air date, and plot details.
</think>
<tool_call>
[
  {
    "function": {
      "name": "visit",
      "arguments": {
        "url": "https://www.moviemistakes.com/tv4916/episode23240/revealing",
        "goal": "Find details about the episode including the title, season, episode number, air date, and plot summary"
      }
    }
  }
]
</tool_call>"
\end{AssistantTurn}

\begin{UserTurn}
user:"Observation from visit:
The useful information in https://www.moviemistakes.com/tv4916/episode23240/revealing for user goal Find details about the episode including the title, season, episode number, air date, and plot summary as follows:

Evidence in page:
The episode in question is titled 'Post Nasal Dick,' which is explicitly mentioned in the URL and heading: 'Post Nasal Dick - S1-E2'. It is part of Season 1, Episode 2 of '3rd Rock from the Sun.' The episode features a revealing mistake involving Harry, who put his feet in a pot of warm chicken soup, but his socks later appear dry.
"
\end{UserTurn}


\begin{AssistantTurn}
assistant:"
<think>
I now have all the confirmation I need.
...
</think>
<answer>
TV Series: 3rd Rock from the Sun
Season: 1
Episode Number: 2
Episode Title: "Post-Nasal Dick"
</answer>"
\end{AssistantTurn}

\section{Released Artifacts and Limitations}
\label{app:release}

We release the evaluation code, tool-use implementation, the 9,221 PrimeSeeker
SFT questions and trajectories, and the construction-verified RL questions with
their reference evidence skeletons. The release also contains source provenance,
the \(\operatorname{Extract}_{D}\) and \(\operatorname{Extract}_{k}\) prompts,
the reference-step auditor, and the evaluation configuration. The exact
upstream prompts used to expand anchor structures and verbalize questions are
not included. Models and data are available at
\url{https://github.com/PengLinzhi/PrimeSeeker}.

Several additional limitations should be considered.

\begin{itemize}
    \item \textbf{Web non-stationarity.}
    Search results and webpage contents may change over time. Reproducing an
    individual trajectory may therefore yield observations different from those
    collected during data construction or evaluation.

    \item \textbf{Evaluator dependence.}
    Benchmark accuracy depends partly on the behavior of
    the answer evaluator. We release the evaluator prompt and code to
    make this source of variation explicit.

    \item \textbf{Evidence-skeleton coverage.}
    \(D_q\) records a compact answer-supporting structure, but it is not intended
    to enumerate every valid search path or every supporting webpage.

    \item \textbf{Open-ended candidate exploration.}
    Latent anchor resolution provides a local formulation of candidate-space
    reduction, but difficult tasks with sparse clues and extremely large
    candidate spaces may still require broader exploration.

    \item \textbf{Model dependence.}
    Closed-book filtering, trajectory generation, and evidence guidance depend
    on the capabilities and search styles of the selected expert models.
\end{itemize}

\end{document}